\documentclass[10pt,twocolumn,letterpaper]{article}

\usepackage{iccv}
\usepackage{times}
\usepackage{epsfig}
\usepackage{graphicx}
\usepackage{amsmath}
\usepackage{amssymb}

\usepackage[pagebackref=true,breaklinks=true,letterpaper=true,colorlinks,bookmarks=false]{hyperref}
\usepackage{subcaption}
\usepackage{multirow}

\iccvfinalcopy 

\def\iccvPaperID{44} 

\ificcvfinal\fi

\newcommand{\keywords}[1]{\textbf{\textit{Keywords---}} #1}

\usepackage{dsfont}
\newcommand{\R}{\mathds{R}}%

\begin{document}

\title{FArMARe: a Furniture-Aware Multi-task methodology for Recommending Apartments based on the user interests}

\author{Ali Abdari\\
University of Udine\\
Via delle Scienze, 206, Udine, Italy\\
University of Naples Federico II\\
C.so Umberto I, 40, Napoli, Italy\\
{\tt\small abdari.ali@spes.uniud.it}
\and
Alex Falcon\\
University of Udine\\
Via delle Scienze, 206, Udine, Italy\\
{\tt\small falcon.alex@spes.uniud.it}
\and
Giuseppe Serra\\
University of Udine\\
Via delle Scienze, 206, Udine, Italy\\
{\tt\small giuseppe.serra@uniud.it}
}

\maketitle
\ificcvfinal\thispagestyle{empty}\fi

\begin{abstract}
   Nowadays, many people frequently have to search for new accommodation options. Searching for a suitable apartment is a time-consuming process, especially because visiting them is often mandatory to assess the truthfulness of the advertisements found on the Web. While this process could be alleviated by visiting the apartments in the metaverse, the Web-based recommendation platforms are not suitable for the task. To address this shortcoming, in this paper, we define a new problem called text-to-apartment recommendation, which requires ranking the apartments based on their relevance to a textual query expressing the user's interests. To tackle this problem, we introduce FArMARe, a multi-task approach that supports cross-modal contrastive training with a furniture-aware objective. Since public datasets related to indoor scenes do not contain detailed descriptions of the furniture, we collect and annotate a dataset comprising more than 6000 apartments. A thorough experimentation with three different methods and two raw feature extraction procedures reveals the effectiveness of FArMARe in dealing with the problem at hand.
\end{abstract}


\section{Introduction}

\begin{figure}
    \centering
    \includegraphics[width=\linewidth,trim=1cm 2.5cm 10.5cm 0.75cm,clip]{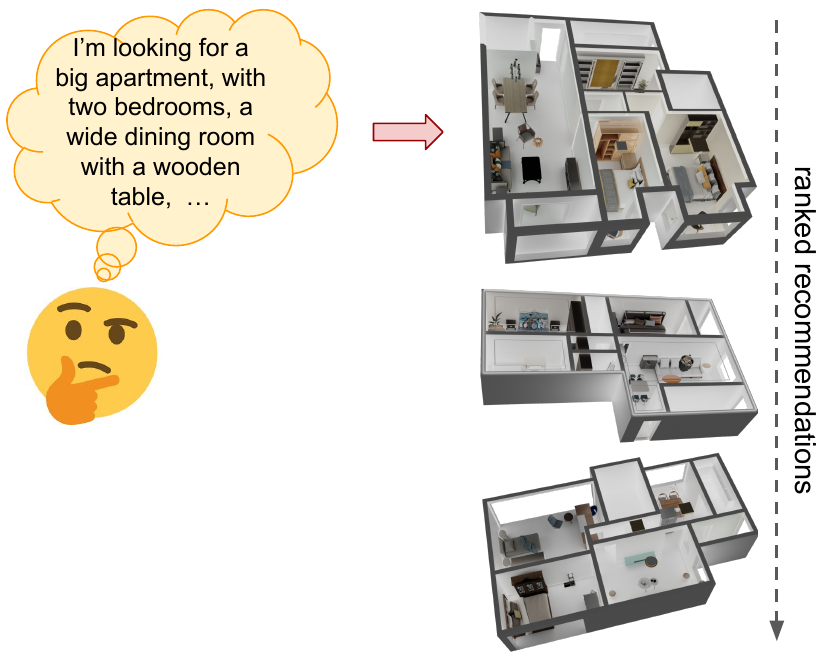}
    \caption{The proposed problem: text-to-apartment recommendation requires ranking all the apartments to recommend those which are most suitable based on the user interests.}
    \label{fig:problem_figure}
\end{figure}

Nowadays, it is common to move across different cities or nations every few years when looking for commercial or educational opportunities. This situation leads to an even more difficult problem, which is finding a new home. To do so, one has to read hundreds or thousands of different advertisements which often highlight the positive aspects of the house, apartment, flat, etc\footnote{In the rest of the paper, we will use apartments to cover these slightly different situations.}, while neglecting the negative ones (e.g., unappealing due to mismatched styles, gloomy due to unsightly furniture and colors, etc). Moreover, even when selecting a subset of the apartments and visiting only those, the process is incredibly time-consuming and stressful, while also becoming more costly and environmentally unfriendly due to the need to physically move to the location. Therefore, visiting digital twins of the apartments in the metaverse would be a more favorable and ideal solution. However, we argue that the dedicated Web applications which are commonly used to ease the search process across all the advertisements, do not represent a good fit. In fact, it is common for these applications to identify a set of properties (e.g., availability, size, price, number of rooms, number of bathrooms, etc) to obtain a filtered list of viable candidates. Yet, these approaches do not allow for a comprehensive analysis of the apartment, as they enable specifying some \textit{properties} but many \textit{unstated user requirements} are left out. Therefore, to truly support the user in the search for an optimal apartment through metaverse exploration, there is a need for an automatic solution which both analyzes the contents of the metaverse scenario and the users' interests to find fewer recommendations that the user can comfortably visit.



The proposed problem, which we name as ``\textbf{text-to-apartment recommendation}'', requires to rank a set of apartments based on their relevance to a given user query, as shown in Figure \ref{fig:problem_figure}. Interestingly, neither this problem nor its more general version related to retrieving 3D scenes based on textual queries, have been studied in the literature. The closest problem which was previously addressed consists in ranking 3d scenes based on 2d images used as queries \cite{abdul2018shrec,abdul2019shrec}. However, such a methodology leads to possible limitations. First, the user needs to have a visualization of the desired apartment, which may be difficult to have, while also limiting the flexibility of the approach (e.g., if the user sketches a floorplan, then it may be difficult to retrieve those apartments which have the same furniture and structural details, yet following a different layout). Second, it also limits the usability from the users' point of view, since describing the desired features in an apartment may be easier than sketching it, while also allowing for more flexibility both with respect to floorplan layout (e.g., by not mentioning the position of the furniture) and to the details which are left out in the query (e.g., if the user does not mention a need for a microwave oven, the retrieved apartments could include both those equipping it and those who do not). The proposed problem, which will be formally described in Section \ref{sec:problem} along its main challenges, aims at allowing the users to describe their interests with free-form text, since this may represent a more comfortable solution, while also avoiding the limitations related to the more traditional Web-based approaches. 

How to tackle this problem? We consider two main factors to support our design choices for a text-to-apartment model. First, the metaverse apartments could be seen as 3d scenes; however, the advertisements for them commonly contain several photos to capture the environment and the furniture available. Second, in recent years there have been impressive advancements in vision and language understanding \cite{radford2021learning}, enabling several multimodal applications and allowing for cross-modal interactions. Therefore, we model our metaverse apartments as a set of images, annotate each apartment with very detailed descriptions, and use a CLIP-based approach to model the cross-modal relations and enable ranking. In particular, we implement several solutions based on deep learning, both inspired by recent literature and from past literature on related topics. The final methodology we propose, which we called FArMARe (Furniture-AwaRe Multi-task methodology for Apartment REcommendation), consists of a multi-task learning approach based on two sub-tasks. First, a contrastive learning objective requires the model to learn cross-modal relations helpful for ranking. Second, a furniture-based classification objective requires the model to more precisely identify the furniture displayed in the images. To experimentally validate our methodology, we collect a benchmark dataset consisting of around 6000 apartments, each of which is annotated by detailed descriptions of its furniture. This is because, as described below, the publicly available datasets do not offer both photo-like indoor pictures and detailed textual descriptions, which are required in our setting.


In summary, the following are the main contributions of this paper:
\begin{itemize}
    \item We introduce the problem of text-to-apartment recommendation, and collect and annotate a benchmark dataset of more than 6000 apartments. 
    \item We design a multi-task learning methodology, called FArMARe, to tackle the problem by introducing a furniture-aware training objective.
    \item We compare the effectiveness of our method with several baselines, inspired by previous research works on similar topics, and show its effectiveness, achieving an improvement of +1.1\% R@1 and +2.0\% R@5.
\end{itemize}

\begin{figure*}[t]
    \centering
    \includegraphics[width=\linewidth]{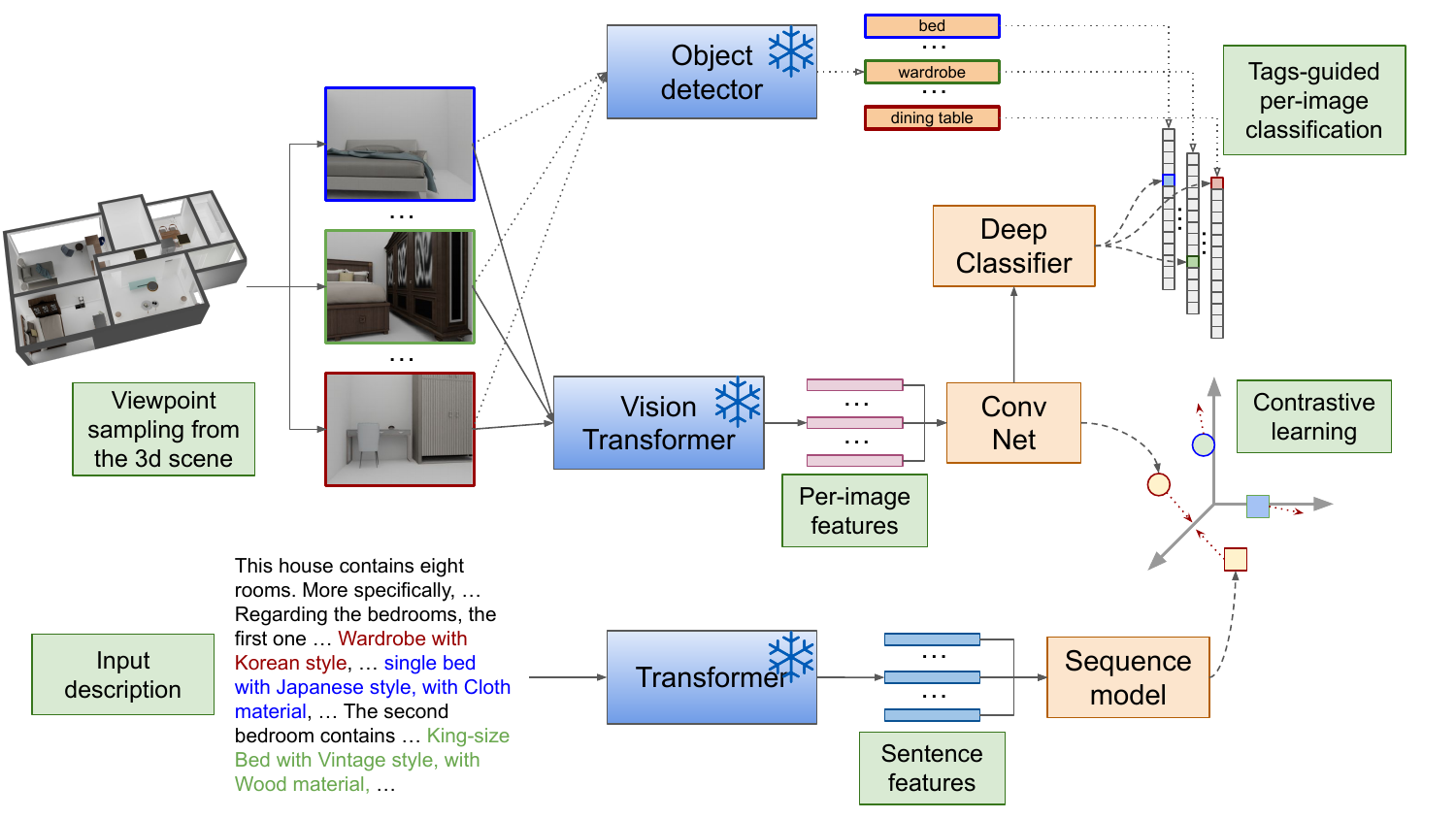}
    \caption{Overview of the proposed methodology. Details in Section \ref{sec:method}.}
    \label{fig:proposed_method}
\end{figure*}

In Section \ref{sec:problem} we describe the proposed text-to-apartment recommendation problem and its challenges. Then, Section \ref{sec:related_work} introduces the scientific literature related to our problem. The proposed methodology is described in Section \ref{sec:method}, whereas a thorough experimental setting is followed in Section \ref{sec:experiments}, and several limitations and future directions are highlighted in Section \ref{sec:discussion_limitations}. Finally, Section \ref{sec:conclusions} concludes the manuscript.

\section{Text-to-apartment recommendation: definition and challenges\label{sec:problem}}
The proposed problem, text-to-apartment recommendation, requires understanding the contents of the 3d scenario representing the apartment, analyzing the user interests, and then connecting them. If the latter are formalized through \textit{free-form} natural language, then the problem can be formally defined as follows. Given a set $\mathcal{A}$ of apartments, $\mathcal{A} = \{a_1, \dots, a_N\}$, each annotated by a description $d_i$, the objective is to learn two functions, $f$ and $g$, such that the apartment $a_i$ is the first retrieved if the query is $d_i$, i.e., $\forall j \ne i, sim(f(a_i), g(d_i)) \ge sim(f(a_i), g(d_j))$, where $sim$ is a similarity metric. In our setting, inspired by the typical structure of the advertisements, each apartment $a_i$ is depicted with a finite set of images.

The proposed problem could be considered similar to other instances of multimedia retrieval, such as text-to-image retrieval \cite{radford2021learning,sangkloy2022sketch}. However, two main intertwined challenges remain. First, there are many details of the apartment which need to be considered, both structural, e.g., how many rooms it has, the presence of separate living and dining rooms, etc, and furniture-related, such as all the modern appliances in the kitchen (fridge, microwave oven, etc). All these details can be inferred from visual analysis, and assessing their relevance to the user interests is very challenging both due to their amount and their different granularity. A second challenge is given by the complexity of modeling all these details into the user interests. In fact, by packing everything into a description, it may end up in very long sentences, made of several hundreds of tokens, which may make it difficult to process and understand. These two problems entail a complex task involving both scene understanding capabilities and cross-modal understanding.


\section{Related work\label{sec:related_work}}
There are several fields which are related to the proposed text-to-apartment recommendation problem. First of all, Section \ref{sec:rw_houses} will discuss the available datasets about apartments, and highlight the motivations which led us to collect a new dataset. Then, Section \ref{sec:rw_scene} will cover recent advancements in scene and language understanding. Finally, in Section \ref{sec:rw_retrieval} we explore the literature on the topic of cross-modal retrieval and contextualize our work into it.

\subsection{About the dataset options for apartments\label{sec:rw_houses}}

The proposed problem involves reasoning on both apartments and textual data. In Table \ref{tab:datasets_characteristics} we summarize several characteristics of the publicly available datasets related to apartments or, more in general, house-like environments. Overall, we looked for a large-scale publicly available dataset containing indoor visual information (e.g., multiple photos or a 3d reconstruction), accompanied by descriptions of the contents, furniture, etc. However, most of the previously published datasets present some shortcomings which limit their applicability to our use case. Houses3k \cite{peralta2020next}, HouseExpo \cite{li2020houseexpo}, and Text-to-3dHouseModel \cite{chen2020intelligent} offer only floorplan representation of the houses, which may not effectively capture appliances and other furniture, thus limiting their usefulness for recommending an apartment. FutureHouse \cite{li2022phyir}, Matterport3d \cite{chang2017matterport3d}, and Habitat-Matterport3d \cite{ramakrishnan2021hm3d} present high-quality panoramic, 3d scenes, images, or other visual formats, yet no language annotations are available. Closest to our work and task are SHREC 2018/2019 \cite{abdul2018shrec,abdul2019shrec,yuan2020comparison}, House3d \cite{wu2018building}, and Rent3d \cite{liu2015rent3d}. However, both Rent3d and SHREC lack textual annotations, whereas House3d has some categorical information but it is not publicly available. Therefore, we decided to collect a new dataset, starting from the realistic rooms available in 3d-FRONT \cite{fu20213d} and creating textual descriptions by aggregating the object, style, theme, and material information available in the dataset (details in Section \ref{exp:dataset}). Noteworthily, this allows for learning to identify the furniture and appliances which are available, whereas the annotations in Text-to-3dHouseModel only described the type of rooms and their location within the house (i.e., not ``detailed'' with respect to the proposed task).

\begin{table*}[t]
    \centering
    \begin{tabular}{|c|ccc|} \hline
        Dataset & Data format & Textual annotations & Scale \\ \hline
        Houses3k \cite{peralta2020next} & Outdoor/exterior & N. A. & Large  \\
        HouseExpo \cite{li2020houseexpo} & Floorplans & N. A. & Very large  \\
        Text-to-3dHouseModel \cite{chen2020intelligent} & Floorplans & Descriptions & Large  \\
        FutureHouse \cite{li2022phyir} & Indoor (panoramic) & N. A. & Very large  \\
        Matterport3d \cite{chang2017matterport3d} & Indoor & N. A. & Large  \\
        HM3d \cite{ramakrishnan2021hm3d} & Indoor & N. A. & Large  \\
        SHREC 2018/2019 \cite{abdul2018shrec,abdul2019shrec} & Indoor & N. A. & Large  \\
        House3d \cite{wu2018building} & Indoor & Category per object/room & Large  \\
        Rent3d \cite{liu2015rent3d} & Floorplans and indoor & N. A. & Small  \\ \hline 
        \textbf{Ours} & \textbf{Indoor} & \textbf{Detailed descriptions} & \textbf{Large}  \\ \hline 
    \end{tabular}
    \caption{Comparison of several datasets from the literature on house models. ``N. A.'' stands for ``Not Available''. Discussion in Section \ref{sec:rw_houses}.}
    \label{tab:datasets_characteristics}
\end{table*}




\subsection{Scene understanding and language\label{sec:rw_scene}}
Early works on scene understanding focused on analyzing the contents of 2d scenes and linking them to textual annotations, for instance when trying to automatize the generation of coherent descriptions \cite{lin2015generating}, providing a correct answer for content-related questions \cite{das2018embodied}, or dealing with text-guided indoor navigation \cite{anderson2018vision}. Compared to the 2d counterpart, similar tasks are far less explored on 3d scenes. In the previous works, the focus was on associating language and single object instances. Chen et al. learned a joint 3d-text embedding space for text-guided generation purposes \cite{chen2019text2shape}. Achlioptas et al. focused on a similar objective, although working at a much finer granularity, learning to distinguish between very fine-grained details of the 3d objects \cite{achlioptas2019shapeglot}. Then, the research focused on more detailed descriptions, containing multiple objects and, possibly, rooms at the same time. This led to the creation of challenging tasks involving the identification of precise instances of an object among many distractors \cite{achlioptas2020referit3d,chen2020scanrefer} and, very recently, the automatic captioning of 3d scenes \cite{chen2021scan2cap}.

The task we propose in this paper is highly related to these advances. Nonetheless, its novelty and importance is highly motivating. First, current research fields are mostly related to localizing or generating objects, which are fundamental tasks yet they may not be as useful when trying to compare complex scenes and assess their relevance to the user interests. Second, the direct implications which our task can have both on businesses related to the metaverse, and on a more environmental-friendly approach to visiting apartments while looking for new accommodation.

\subsection{Multimedia retrieval\label{sec:rw_retrieval}}
Finding multimedia content which is relevant to the user interests among many hundreds of thousands or even millions of examples is a difficult problem which is getting more and more attention from the research community \cite{girdhar2023imagebind,radford2021learning}. The users commonly describe their interests by means of textual queries, which are then automatically mapped into a joint embedding space where another type of multimedia content is also mapped, e.g., images \cite{radford2021learning,sangkloy2022sketch} or videos \cite{falcon2022feature,girdhar2023imagebind}. Recently, even more modalities were taken into consideration when building these spaces, e.g., by considering simultaneously text, image/video, and audio \cite{bain2021frozen,shvetsova2022everything}. The problem we are facing in this paper can be seen as retrieving apartments (i.e., scenes) by means of textual queries, as described in Section \ref{sec:problem}. This problem is interesting both for the challenges which we highlighted before, for its practical use cases, and also for its novelty. In fact, to the best of our knowledge, there are no previous works addressing the problem of \textit{retrieving apartments}, or even more general 3d scenes, \textit{by means of textual queries}. The only line of work about retrieving 3d scenes involves the use of images or sketches as queries \cite{abdul2018shrec,abdul2019shrec,yuan2019sketch}. However, the queries are visual and not textual, therefore limiting the usability of those approaches in our setting. Other works on `scene retrieval' focus on locating the objects in the scene \cite{nguyen2020robot} or the text shown inside them \cite{wang2021scene,wen2023visual}.


\section{Proposed method: FArMARe\label{sec:method}}
An overview of the proposed methodology, called Furniture-AwaRe Multi-task methodology for Apartment REcommendation (or FArMARe), is shown in Figure \ref{fig:proposed_method}. The textual and visual data are processed by two independent branches, as typically done in cross-modal retrieval \cite{cao2022image,lei2021less,radford2021learning,zhu2023deep}. In the following sections, we describe each of these branches separately, and how they are combined together.

\subsection{Visual data}

Starting from the 3d scene, multiple viewpoints are sampled, the amount of which depends on a subset of furniture categories present in the scene, as done in \cite{dahnert2021panoptic}. Each of the images is then processed by a pretrained Vision Transformer \cite{dosovitskiy2020image} to obtain a set of image descriptors. Then, our multi-task approach requires solving two tasks: ranking, and classification. For the first task, we use a simple network to learn a function which maps each image descriptor $x_i$ into the joint embedding space:
\begin{equation}\label{eq:conv1d}
    \hat{x}_i = Conv1d(x_i)
\end{equation}
\begin{equation}
    a_i = ReLU(\hat{x}_i W_c + b_c)
\end{equation}
\noindent where $W_c$ and $b_c$ are trainable parameters. At the same time, $\hat{x}_i$ is also processed by the network solving the second task, i.e., classification. The classifier learns to predict a class $t_i$ for each of the images starting from the shared representation $\hat{x}_i$. To do so, it uses the following equations: 

\begin{eqnarray}\label{eq:classifier}
    r_1 = ReLU(\hat{x}_i W_{c_1} + b_{c_1}) \nonumber\\
    r_2 = \delta(ReLU(r_1 W_{c_2} + b_{c_2})) \nonumber\\
    r_3 = ReLU(r_2 W_{c_3} + b_{c_3}) \\
    r_4 = ReLU(r_3 W_{c_4} + b_{c_4}) \nonumber\\
    p_i = softmax(r_4 W_{c_5} + b_{c_5}) \nonumber
\end{eqnarray}

\noindent where $W_*$ and $b_*$ are trainable parameters, and $\delta$ applies dropout with probability 0.1. The supervision for the classification is provided by automatically recognized furniture tags, leading to furniture-awareness in the training objective and in the features extracted in the intermediate layers. These tags are obtained by YOLOv8 \footnote{https://yolov8.com/} which we finetune on a dataset of furniture images \cite{furnituredataset} considering 25 different classes, e.g., `bed', `shelf', etc. In this way, the model needs to extract more general, furniture-aware features which are useful to solve both tasks simultaneously.

\subsection{Textual data}

As mentioned before, describing all the important aspects of an apartment may result in a very long description. Although the recent advancements in natural language processing made it easier to process and understand their contents \cite{bulatov2023scaling,chen2023extending}, very long sentences may still pose a problem \cite{shaham2022scrolls}. To this end, we opted for the following approach. First of all, the apartment description is split into several sentences by splitting at the periods. Then, a pretrained Transformer \cite{vaswani2017attention} is used to extract several sentence-level representations, after which a bidirectional GRU is trained contrastively with the visual data. The final representation for the apartment description, $d_i$, is obtained by taking the mean of the final hidden state of the bidirectional GRU.

\subsection{Loss function}

The proposed approach, FArMARe, simultaneously learns to solve two tasks. For the first task, i.e., ranking, we use the triplet loss \cite{schroff2015facenet} which is commonly used in cross-modal retrieval settings \cite{bhunia2022adaptive,falcon2022feature,wang2023learning}. Given a batch of $N$ apartments, $a_1, \dots, a_N$, and their descriptions $d_1, \dots, d_N$, it can be defined as follows:
\begin{align*}
    l_r(a, p, n) &= \max(0, \Delta + s(a, n) - s(a, p)) \\
    \mathcal{L}_{r} = \frac{1}{2 \cdot N \cdot (N-1)} &\big( \sum_i \sum_{j \ne i} l_r(a_i, d_i, d_j) \\
    &\quad + \sum_i \sum_{j \ne i} l_r(d_i, a_i, a_j) \big)
\end{align*}

For the second task, i.e., classification, cross-entropy is used. It is defined as follows:
\begin{equation*}
    \mathcal{L}_c = - \frac{1}{N} \sum_{j=1}^{N} t_{i, j} \log(p_{i, j})
\end{equation*}
\noindent where $p_{i, j}$ is the softmax-normalized logit for the $j$-th class, and the tag $t_i$ is seen as a one-hot representation where $t_{i, i} = 1$ and $t_{i, j} = 0$ for $i \ne j$. The final loss function is defined as:
\begin{equation*}
    \mathcal{L} = \frac{1}{2} \big( \mathcal{L}_c + \mathcal{L}_r \big)
\end{equation*}

\section{Experimental results\label{sec:experiments}}

The source code used in this paper for creating the dataset and training the models can be found on this     \href{https://github.com/aliabdari/FArMARe}{Github}.

\subsection{Analysis of the dataset\label{exp:dataset}}
\begin{figure}
    \centering
    \includegraphics[width=\linewidth]{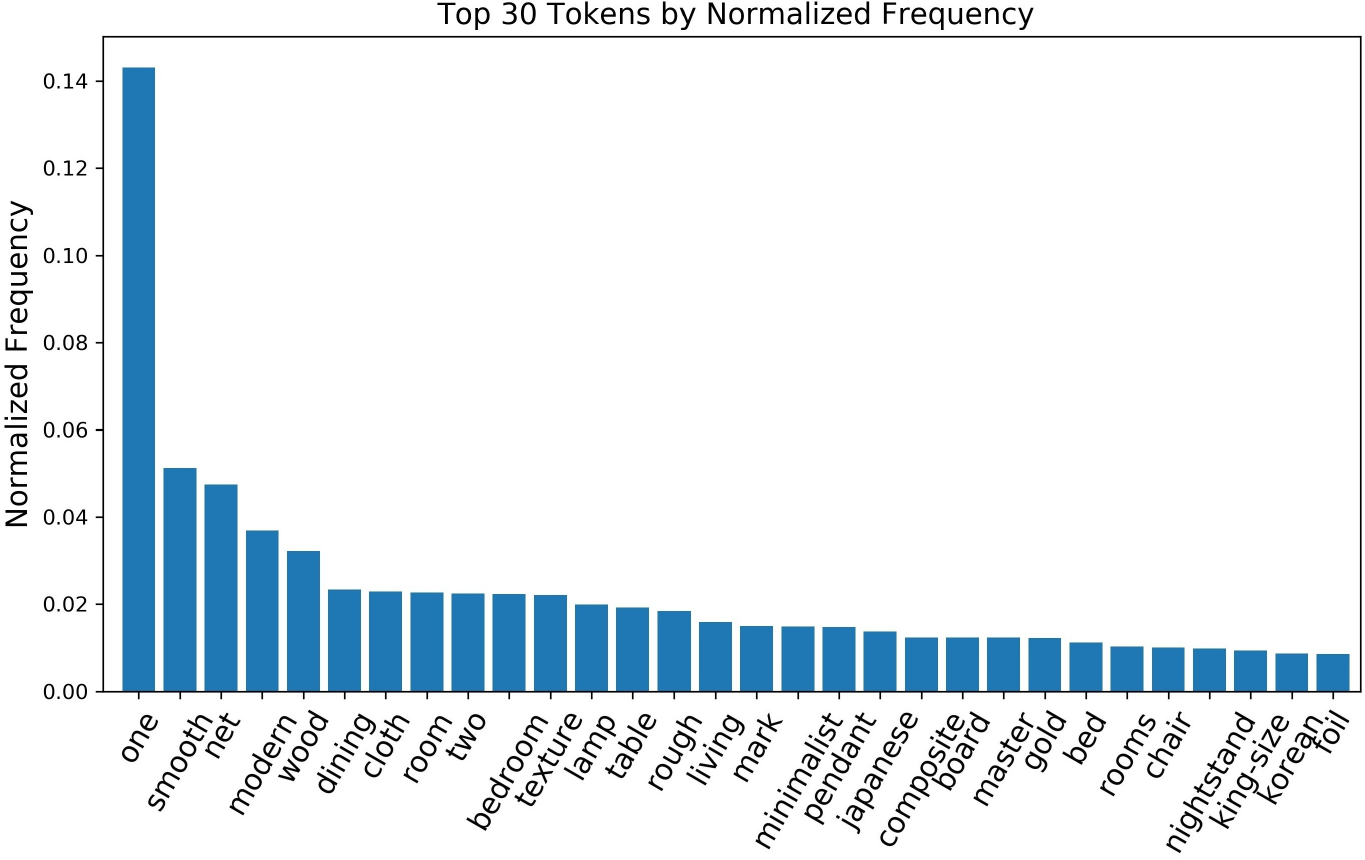}
    \caption{Top 30 most common tokens in the collected dataset. Discussion in Section \ref{exp:dataset}.}
    \label{fig:top30tokens}
\end{figure}

The dataset used in this paper consists of 6081 indoor apartments which we gathered from 3d-FRONT \cite{fu20213d}. Each apartment is annotated with a textual description which we collected by using the object categories, style, theme, and material annotations associated with each piece of furniture. On average, the descriptions have 319 words, ranging from a minimum of 19 and a maximum of 1905, making them very long; looking at the sentences, the average reduces to 16.4 (2 to 93), making them much easier to process. The words encountered in the annotations are highly specific to the task at hand, and describe several styles for the furniture (e.g., ``Japanese'' and ``minimalist''), themes (e.g., ``smooth'' and ``net''), and materials (e.g., ``wooden'' and ``composite''). In Figure \ref{fig:top30tokens} we report the normalized frequency of the top 30 most common tokens, after the removal of stop-words and some words we use to create more human-like descriptions (e.g., ``moreover'', ``additionally'', etc). It can be seen that, apart from ``one'' which is common, the other tokens are not so frequent, making the retrieval task challenging, due to more difficulties in capturing relationships between words.

\subsection{Experimental setting\label{exp:exp_setting}}
We use two evaluation metrics to assess the performance of the approaches under analysis. The first metric, \textbf{recall@k} (or R@k), measures a model's ability to successfully retrieve the groundtruth information among the top k recommendations. Moreover, we also include the Rsum, which consists of the sum of R@1, R@5, and R@10 both for text-to-apartment and apartment-to-text. The second metric is the \textbf{median rank}, which reveals the position of the groundtruth in the ranked list. A lower median rank value indicates that the groundtruth is more likely to be found at higher positions, demonstrating better ranking accuracy.

In the following experiments, we both explore the use of a jointly trained vision and language backbone, and separately trained backbones. For the former, we consider a Vision Transformer, ViT-B-32 \cite{dosovitskiy2020image}, and a 12-layer Transformer jointly trained via CLIP. The pretraining is done on LAION-2B (a subset of \cite{schuhmann2022laion}). For the latter, we consider an ImageNet-pretrained ResNet-152 \cite{he2016deep} and a pretrained BERT \cite{devlin2019google}. Is it important to note that in the non-learning baseline, only the jointly trained backbone is considered: in fact, it would not be possible to compute meaningful similarity scores by using ResNet and BERT, since the two share completely unrelated embedding spaces.

Here, we provide a brief description of the baseline methods. Overall, all the learning-based methods use sentence-level textual features followed by a bidirectional GRU for the text data modeling.

\textbf{Non-learning baseline (NLB).} In SHREC 2018/2019, which is the closest work to ours (see Section \ref{sec:rw_houses}), non-learning based solutions were also proposed. The best-performing one used a pretrained VGG to extract features for both input data (i.e., the 2d query, and the 2d viewpoints of the 3d scene) and then computed the similarity matrix to perform the ranking. In our setting, this approach can not be used as is: in fact, our queries are textual and not visual, therefore imposing a domain gap too big to obtain a decent result. To address this and propose a non-learning baseline, we use a pretrained model jointly trained with CLIP \cite{radford2021learning} to extract both textual and visual features, separately pool them, and then compute the similarity matrix via cosine similarity. The model uses ViT-B-32 \cite{dosovitskiy2020image} for the visual backbone, and a 12-layer Transformer \cite{vaswani2017attention} for the textual one.

\textbf{Average-pool based, FC network (AFN).} To aggregate the visual descriptors $x_i$ obtained from the backbone, this method simply averages them, obtaining $r_1 \in \R^{1 \times F_V}$. Then, the following network is used to learn the final descriptor for the apartment, $a_i$:
\begin{eqnarray*}
    r_2 = BN(\delta_1(ReLU(W_{f_1} r_1 + b_{f_1}))) \\
    r_3 = BN(\delta_2(ReLU(W_{f_2} r_2 + b_{f_2}))) \\
    a_i = W_{f_3} r_3 + b_{f_3}
\end{eqnarray*}
\noindent where $W_*, b_*$ are trainable weights and biases, $\delta_1(\cdot)$ and $\delta_2(\cdot)$ represent the dropout operator with probability 0.20 and 0.10, respectively. $BN(\cdot)$ identifies the use of Batch Normalization \cite{ioffe2015batch}.

\textbf{Conv1d network (CNV).} Different from the previous method, the visual descriptors are not initially pooled. Instead, we use a simple network consisting of a 1d convolutional network to reduce the number of features, a ReLU activation function, and finally, a fully-connected layer to learn the final representation for the apartment.

\subsection{Implementation details}

For the 1d convolution used in Eq. \ref{eq:conv1d} we use 512 output channels, stride one, and the kernel size of five. The classifier in Eq. \ref{eq:classifier} reduces the dimension by half in each layer, therefore going from 512 to 64. Then, we use the following 25 predefined classes: Bed, Cabinet, Carpet, Ceramic floor, Chair, Closet, Cupboard, Curtains, Dining Table, Door, Frame, Futec frame, Futec tiles, Gypsum Board, Lamp, Nightstand, Shelf, Sideboard, Sofa, TV stand, Table, Transparent Closet, Wall Panel, Window, and Wooden floor. Moreover, we used a 26th class for those tags whose confidence was lower than 10\%. To reduce the domain gap between the pretraining data and the data at hand, we finetuned YOLOv8 on 8000 images of furniture \cite{furnituredataset}.

We use PyTorch 1.13.1 for the implementation and run all the experiments on a machine using an RTX A5000 GPU, 16 GB of RAM, and an Intel Xeon E5-1620. The training lasts 50 epochs with a batch size of 64. The optimizer is Adam and the learning rate starts from .008 and is decayed by a factor of 25\% after 27 epochs.

\subsection{Reasoning about the input data\label{exp:input_format}}
The first two research questions we aim at answering are the following: (1) In the proposed setting, is it better to have the visual and textual backbones jointly or separately trained? (2) Considering that the descriptions feature hundreds of tokens, is it better if we extract the textual features at the token- or at the sentence-level?

To answer both questions, we perform the following experiment using the \textit{AFN} method. For the jointly trained backbones, we consider ViT-B-32 and a Transformer trained with CLIP; for the other question, we use ImageNet-pretrained ResNet-152 \cite{he2016deep} and BERT \cite{devlin2019google} to separately extract visual and textual features, respectively. We evaluate the four combinations (separate or joint features, token, or sentence level) and report the results in Table \ref{tab:feature_and_input_format}. To answer the first question, it can be seen that the features extracted from a jointly trained vision and language backbone lead to far better performance, achieving around 35.4\% and 48.9\% text-to-apartment R@10 at the token- and sentence-level respectively. These results represent a margin of +30.8\% and +32.0\%, respectively, compared to the separately trained features. For the second question, it can be seen that working on sentences makes the feature extraction process more effective, obtaining 16.9\% R@10 using ResNet and BERT (+12.3\% than the token-level), and 48.9\% R@10 using CLIP (+13.5\%).

\begin{table}[t]
    \centering
    \begin{tabular}{c|c|c|}
        \textit{Visual $\vert$ Textual} & Token-level & Sentence-level \\ \hline
        ResNet $\vert$ BERT &4.6  &16.9  \\ \hline
        CLIP $\vert$ CLIP &35.4  &48.9  \\ \hline
    \end{tabular}
    \caption{Analysis of the performance (text-to-apartment R@10) when changing the input format for the textual annotations (long list of token-level features or shorter list of sentence-level features) and type of features (separately learned via ResNet-152 and BERT or jointly learned via CLIP). Discussion in Section \ref{exp:input_format}.}
    \label{tab:feature_and_input_format}
\end{table}

\subsection{Comparison between the baseline methods\label{exp:comparison}}
In Table \ref{tab:baseline_comparison} we report a comprehensive quantitative analysis of the performance obtained by the three models we considered in our study (details in Section \ref{exp:exp_setting}) and by our proposed method. As mentioned in the previous sections, it is not meaningful to use the non-learning baseline with ResNet and BERT, as the two present a wide domain gap between the two single-modal spaces. Then, from the Table, three main results are observed. First, all the methods under analysis achieve better performance when using the jointly trained backbones, as opposed to the separately trained ones, confirming the results obtained in the previous experiment. Second, delaying the fusion of the information obtained from the single viewpoints leads to better results than performing it early, since both CNV (e.g., 63.7 and 225.9 Rsum with ResNet and BERT, and CLIP features respectively) and FArMARe (92.3 and 232.7 Rsum) obtain far better performance than AFN (33.2 and 133.7 Rsum). Finally, the information obtained by the additional task in our multi-task approach leads to a considerable improvement over CNV, obtaining +1.1\%, +2.0\%, and +0.2\% text-to-apartment R@1, R@5, and R@10, and overall +6.8 Rsum. Noteworthily, these results are obtained by performing 5 runs with both of the models, to reduce the influence of randomness.

\begin{table*}[]
    \centering
    \begin{tabular}{c|cccc|cccc|c|}
         & \multicolumn{4}{c|}{Text-to-Apartment} & \multicolumn{4}{|c|}{Apartment-to-Text} & \multirow{2}{*}{Rsum} \\ \cline{2-9}
        Method & R@1 & R@5 & R@10 & MedR & R@1 & R@5 & R@10 & MedR & \\ \hline \hline
        \multicolumn{10}{c|}{ResNet $\vert$ BERT features} \\ \hline
        AFN &1.3  &5.8  &10.4  &101  &1.6  &4.6  &9.5  &104  &33.2  \\
        CNV &4.3  &11.4  &16.6  &93  &2.7  &11.8  &16.9  &95  &63.7  \\ \hline
        \textbf{FArMARe (Ours)} &\textbf{6.9}  &\textbf{17.4}  &\textbf{22.1}  &\textbf{88} & \textbf{7.1}  &\textbf{16.7}  &\textbf{22.1}  &\textbf{89} & \textbf{92.3}  \\ \hline \hline
        \multicolumn{10}{c|}{CLIP features} \\ \hline
        NLB ($\sim$ \cite{abdul2018shrec}) & 0.1  &0.6  &1.3  &413  &0.6  &1.9  &3.7  &339  &8.2  \\
        AFN &10.6  &25.5  &34.2  &29  &8.9  &23.2  &31.3  &31  &133.7  \\
        CNV &23.7  &40.6  &48.9  &\textbf{11}  &23.1  &40.7  &48.9  &12  &225.9  \\ \hline
        \textbf{FArMARe (Ours)} &\textbf{24.8}  &\textbf{42.6}  &\textbf{49.1}  &\textbf{11}  &\textbf{25.7}  &\textbf{41.4}  &\textbf{49.1}  &\textbf{11}  &\textbf{232.7}  \\ \hline
    \end{tabular}
    \caption{Comparison between the baselines under analysis (Section \ref{exp:exp_setting}), and the proposed method, FArMARe (Section \ref{sec:method}). Discussion in Section \ref{exp:comparison}.}
    \label{tab:baseline_comparison}
\end{table*}

\section{Discussion and limitations\label{sec:discussion_limitations}}
\begin{figure*}[t]
    \centering

    \begin{subfigure}{0.3\linewidth}
        \includegraphics[width=\linewidth]{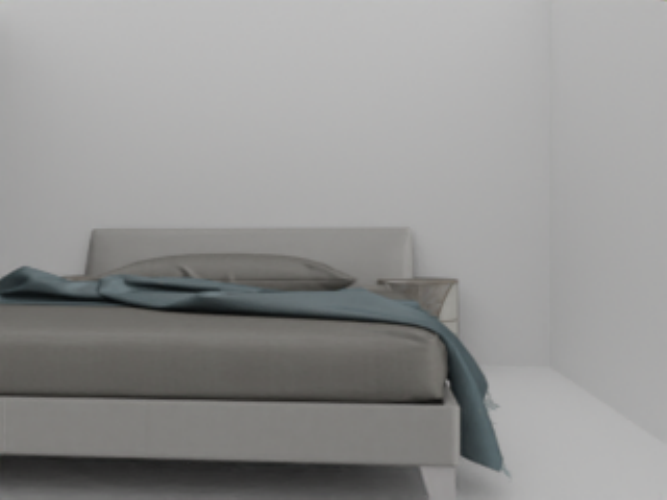}
        \caption{Tag: ``bed''}
    \end{subfigure}
    \hfill
    \begin{subfigure}{0.3\linewidth}
        \includegraphics[width=\linewidth]{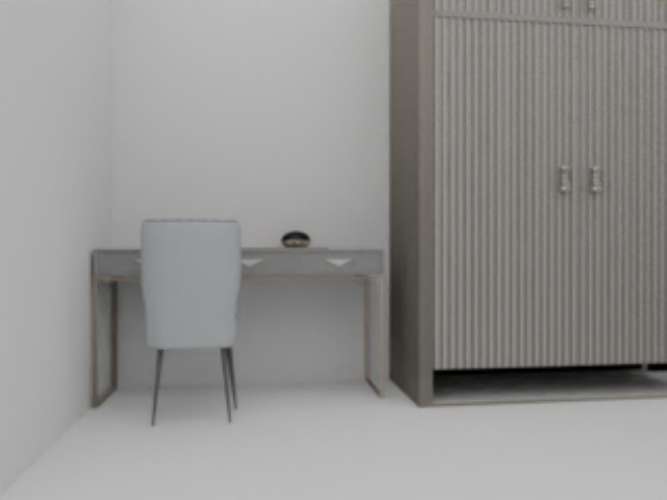}
        \caption{Tag: ``wardrobe''}
    \end{subfigure}
    \hfill
    \begin{subfigure}{0.3\linewidth}
        \includegraphics[width=\linewidth]{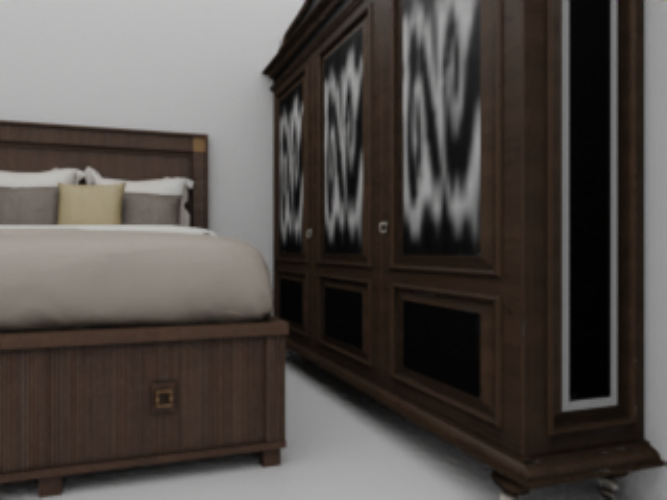}
        \caption{Tag: ``dining table''}
    \end{subfigure}

    \caption{Qualitative examples of the tags obtained from the object detector, including examples when (a) it correctly identifies the predominant furniture; (b) it correctly identifies one of the furniture, but misses the other; (c) it fails. Discussion in Section \ref{sec:discussion_limitations}.}
    \label{fig:limitation_detection}
\end{figure*}

\textbf{About the object detection module.} As mentioned in Section \ref{sec:method}, we are currently extracting the best tag according to the detector. This works well when there is only one main object in the viewpoints (e.g., see Figure \ref{fig:limitation_detection}.a), however when multiple are visible picking only one tag may not be enough (e.g., see Figure \ref{fig:limitation_detection}.b). Moreover, the detection may also fail, leading to incorrect information injected into the visual descriptors (e.g., see Figure \ref{fig:limitation_detection}.c). While overall the proposed method proves to be effective, these limitations may indeed affect its final performance and reliability, leaving much space for improvement. We highlight the following options to address these shortcomings. First, considering that multiple object tags may be relevant for a single image, a different learning strategy could be designed to leverage this aspect (i.e., moving to a multi-label classification problem), while also adopting spatial attention to ground the predictions, making the approach both more explainable and possibly improving the overall performance. Second, in our setting, which consists of photorealistic yet synthetic environments, generating the supervision from synthetic data may help obtain a more precise classifier, which could also lead to better features for the ranking task.

\textbf{About the long descriptions.} In our methodology, we decided to split the descriptions into a sequence of sentences, then to capture the meaning of each of them separately by using a powerful language model, and finally to model their contextual information through a neural sequential model. By dividing the language understanding step into two sub-steps, we are able to achieve higher effectiveness. However, recent advancements showed that very long contexts can be understood by large language models \cite{bulatov2023scaling,chen2023extending}. Therefore, by extending our methodology with these techniques it may be possible to obtain better language understanding even while working at the token-level.

\section{Conclusion\label{sec:conclusions}}
In this paper, we defined the problem of recommending an apartment based on user-defined queries which capture their interests. This novel problem is inspired by the need for applications which are able to analyze the contents of a digital twin of the apartment, representing it in the metaverse, assess whether it is a good match for the user-defined query, and rank the metaverse apartments accordingly. This approach may be highly impactful both on businesses dealing with apartments recommendation, and on the environment, since being able to visit the apartment in the metaverse has a smaller impact than visiting it by physically moving across states.

To realize a model able of solving the problem, we introduced FArMARe, a multi-task methodology which uses both a contrastive framework, to learn how to perform cross-modal ranking, and a furniture-based classification objective, which helps the model extracting more general furniture-aware features. We collected and annotated a dataset of more than 6000 apartments, over which we performed several experiments with three different methods and two different feature extraction procedures, confirming the effectiveness of the proposed method. 

Finally, we discussed the limitations of the model, while also highlighting possible future research directions.

\section*{Acknowledgements}
This work was supported by the Department Strategic Plan (PSD) of the University of Udine–Interdepartmental Project on Artificial Intelligence (2020-25), by the Italian Ministry of University and Research (MUR) Progetti di Ricerca di Rilevante Interesse Nazionale (PRIN) 2022 (project code 2022YTE579) and within the project DM737\_HEU\_voucher\_2b\_FALCON (CUP G25F21003390007), and by TechStar Srl, Italy.

{\small
\bibliographystyle{ieee_fullname}
\bibliography{egbib}
}

\end{document}